\documentclass{article}

\usepackage[nonatbib, preprint]{neurips_2020}

\usepackage[utf8]{inputenc} 
\usepackage[T1]{fontenc}    
\usepackage{hyperref}       
\usepackage{url}            
\usepackage{booktabs}       
\usepackage{amsfonts}       
\usepackage{nicefrac}       
\usepackage{microtype}      
\usepackage{graphicx}
\usepackage{enumitem}
\usepackage{amsmath}
\usepackage{esvect}
\usepackage{xcolor}
\usepackage{enumitem}
\usepackage{siunitx}
\usepackage{mathtools}
\usepackage{caption}
\usepackage{diagbox}
\usepackage{multirow}
\usepackage{float}
\usepackage{diagbox}
\usepackage{multirow}
\usepackage{xspace}
\usepackage[title]{appendix}

\setlist{noitemsep}

\long\def\comment#1{}

\newcommand{\varbold}[1]{\textcolor{black}{#1}}

\newcommand{\niv}[1]{\textcolor{black}{#1}}

\newcommand{\ben}[1]{\textcolor{black}{#1}}
\newcommand{\michal}[1]{\textcolor{black}{#1}}

\newcommand{\nivc}[1]{\comment{#1\}}}

\newcommand{\benc}[1]{\comment{#1\}}}

\newcommand{\CC}{\textsc{CC}\xspace}

\usepackage{biblatex}
\vspace{-0.4cm}

\title{From Discrete to Continuous Convolution Layers}
\begin{document}
\author{Assaf Shocher\thanks{equal contribution}
\qquad Ben Feinstein\footnotemark[1]
\qquad Niv Haim\footnotemark[1]
\qquad Michal Irani\\
\small 
Dept. of Computer Science and Applied Math, The Weizmann Institute of Science
}

\maketitle
\vspace{-0.6cm}
\begin{abstract}
\vspace{-0.2cm}
A basic operation in Convolutional Neural Networks (CNNs) is spatial resizing of feature maps. This is done either by strided convolution (donwscaling) or transposed convolution (upscaling). Such operations are limited to a fixed filter moving at predetermined integer steps (strides). Spatial sizes of consecutive layers are related by integer scale factors, predetermined at architectural design, and remain fixed throughout training and inference time. We propose a generalization of the common Conv-layer, from a discrete layer to a Continuous Convolution (CC) Layer. CC Layers naturally extend Conv-layers by representing the filter as a learned continuous function over sub-pixel coordinates. This allows learnable and principled resizing of feature maps, to any size, dynamically and consistently across scales. 
Once trained, the CC layer can be used to output any scale/size chosen at inference time. The scale  can be non-integer and differ between the axes. CC gives rise to new freedoms for architectural design, such as dynamic layer shapes at inference time, or gradual architectures where the size changes by a small factor at each layer. This gives rise to many desired CNN properties, new architectural design capabilities, and useful applications. We further show that current Conv-layers suffer from inherent misalignments, which are ameliorated by CC layers.

\end{abstract}
\section{Introduction}

The basic building block of CNNs, the convolution layer (conv-layer), applies a discrete convolution (more precisely, cross-correlation) to an input feature map, with a learned discrete filter.
CNNs commonly employ spatial resizing of their feature maps, either downscaling (e.g., in classification networks) or upscaling (e.g., in generative models and image processing tasks). These resizing operations are either limited to an integer scale factor (achieved by an integer stride), or via unprincipled feature map interpolation. These resizing approaches suffer from inherent problems (elaborated below), which limit the performance and capabilities of current CNNs.


In this paper, we propose a generalization of the standard conv-layer. 
Given a discrete feature map $I$ of 
spatial dimensions  $H$$\times$$W$,
we are interested in a learned conv-layer that outputs a feature map $I'$ of $H'$$\times$$W'$, 
where none of the ratios of the spatial dimensions ($\frac{H}{H'}$,$\frac{H'}{H}$,$\frac{W}{W'}$,$\frac{W'}{W}$)
need to be integer numbers. We show that this can be achieved by modeling a \emph{learned continuous convolution}, which is end-to-end trainable by gradient descent. Furthermore, we show that such continuous modeling  allows for the desired \ben{output size} $H' \times W'$  to be chosen dynamically, \emph{at inference time}. This gives rise to many desired CNN properties, new architectural design capabilities, and useful applications. 

\begin{figure}
    \centering
    \hspace*{-0.5cm} \includegraphics[width=1.05\textwidth]{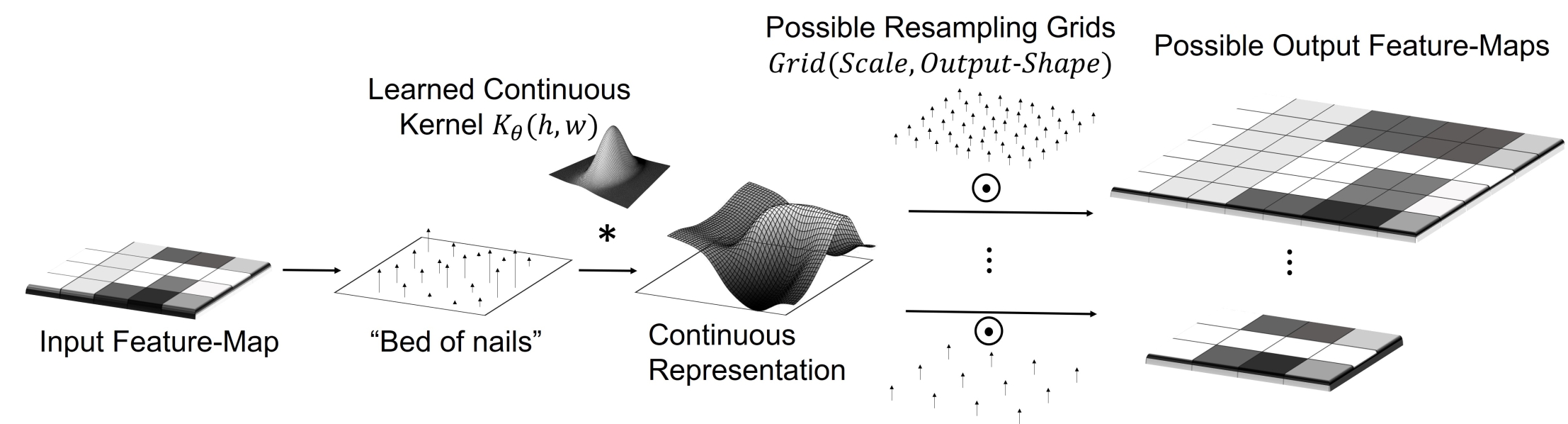}
    \caption{The underlying process of CC.
    }
    \label{fig:underlying}
\end{figure}

Fig.~\ref{fig:underlying} shows the underlying process of a CC-layer, which has two parts: (i) modulating the discrete input with a learned continuous filter to get a continuous intermediate representation; (ii) re-sampling the continuous representation according to \ben{a} re-sampling grid to get the discrete output. The \ben{only} learned part in a CC-layer is the continuous filter. Once trained, the CC-layer can be used to output any scale or shape chosen at inference time, simply \ben{by using a different re-sampling grid}.


To date, the most common approach\niv{es} \ben{for} \ben{feature map resizing} in CNNs \niv{are} the strided-conv or pooling for  downscaling, or the transposed-conv for upscaling. These suffer from inherent limitations: (i)~The stride must be an integer, hence a learned transformation that includes spatial resizing can only be done by an integer (or inverse of an integer) scale-factor. (ii)~Once trained, the filter and stride are fixed and cannot be modified at inference time. (iii)~Strided-convolutions suffer from impaired shift-\ben{equivariance}~\cite{zhang2019shiftinvar, azulay2018deep}. Transposed-convolutions suffer from inherent checkerboard artifacts~\cite{odena2016deconvolution}.

Interpolation-based resizing methods (e.g., linear, nearest neighbor, cubic~\cite{keys1981cubic}), which are mostly used for images, are usually differentiable and can be incorporated in a neural network to resize feature maps. Such resizing is principled; it follows a consistent rule across all scales and sub-pixel/sub-feature positions. 
However, these are not \emph{learned} resizing operations, and furthermore cannot express more general feature transforms, only scaling. We do however draw inspiration from the way these methods are applied to images~\cite{MATLAB:2010}, 
and show how 
they can be replaced by dynamic learnable models.
%


The most closely related work to ours is that of~\cite{wang2018deep}. 
They were the first to coin the term ``Continuous Convolution'', and identified its need for applying CNNs on 3D point-clouds (points scattered irregularly in 3D space; not on a regular grid).  
They used a method similar to ours for calculating convolution weights for points in the irregular 3D cloud. However, modeling continuous convolution for point-clouds is 
a clear necessity, as  there is no notion of ``stride'' in such input data. Moreover, their continuous convolution is not meant to resize and is used for a different goal. Inspired by their work, we generalize the concept of Continuous Convolution of~\cite{wang2018deep} in several ways: (i)~We adapt it to the realm of \emph{regular grids}, where stride/scale is well-defined, yet show that it is useful also here. (ii)~Our continuous convolution can be consistently applied to any choice of scale and output shape, and (iii)~the scale and shape of our  CC-layer output can be determined dynamically at inference time.

It was shown by~\cite{azulay2018deep,zhang2019shiftinvar} that very small shifts to the input data of a CNN (e.g., shifting an input image by 1-2 pixels) can drastically change the output prediction. This lack of shift- \ben{invariance/equivariance} is due to the aliasing induced by strided conv-layers with small kernel support. CC enables more ``gradual-architectures'' (still with small kernel support) that avoid aliasing, thus giving rise to true shift-equivariance.



\textbf{The contributions of our paper are therefore several-folded:}
\vspace*{-0.3cm}
\begin{itemize}[noitemsep,leftmargin=0.7cm]
    \item \emph{\textbf{From Discrete to Continuous CNNs:}} We introduce a generalization of the standard discrete conv-layer, to a continuous CC-layer, 
    in a principled way.
    \item \emph{\textbf{Resize \ben{by} any scale:}} CC is the first learnable layer that can resize feature-maps \ben{by} any scale (integer or non-integer; downscaling/upscaling; can also differ between dimensions).
    \item  \emph{\textbf{Dynamic:}} 
    The desired scale and shape of the CC \ben{output} can be \ben{set} at \ben{inference}-time.
        \item \emph{\textbf{True shift-equivariance:}}  CC enables gradual-downscaling that induces less aliasing.
    \item \emph{\textbf{Resolving input-output misalignments:}} 
    We further show that conv-layers often induce small inherent misalignments, 
    which are ameliorated by CC.
\vspace*{-0.3cm}
\end{itemize}

\section{Overview of the approach}
\label{sec:overview}
Fig.~\ref{fig:underlying} describes the underlying process we aim to model:
given a discrete feature map $I[i,j]$, we are interested in constructing \niv{a} CC-layer that operates on $I$ as an input, and produces a next layer $I'[i',j']$, by modeling a learned continuous convolution between the input and output feature-maps.
Note that the input and output feature-maps  $I$ and $I'$ need \emph{not} be related by an integer  scale factor (or its inverse).
Such \niv{an} action can be realized by \niv{the following steps}: 
\vspace*{0.1cm} \\ 
(i)~Use a ''Bed of Nails'' representation of $I$:
$I_{cont}(h,w) = \sum_{i,j}\delta(h-i, w-j)I[i,j]$.
where $i,j$ are discrete `pixel' (feature) locations in $I$, and $h,w$ are continuous coordinates in the continuous feature-map space. 
For simplicity, {we will refer from here on  to all discrete coordinates inside feature maps by the term \textbf{`pixels'}, and their in-between non-integer coordinates as \textbf{sub-`pixel'} (although we are referring to general network layers; usually not images)}. 
\vspace*{0.1cm} \\
(ii)~Apply a convolution with a \emph{learned} continuous kernel $\mathcal{K}_\theta(h,w)$.
\vspace*{0.1cm} \\
(iii)~Resample the continuous result of the convolution according to the desired shape and scale-factors of $I'$. The resampling grid is purely a function of the shape and scales.
This means that the same continuous kernel $\mathcal{K}_\theta$ is eligible for producing any desired scale-factor for $I'$.

These phases are indicated in Fig.~\ref{fig:underlying}, and can be mathematically formulated as follows:
\begin{align}
    CC\{I\}[\textbf{n}] = \{I_{cont} * \mathcal{K}_{\theta}\}(\textbf{g}_n)
    &=
   \iint\sum_{\textbf{m}}\delta
   \left(\textbf{g}_n-\mathbf{\tau}-\textbf{m}\right)
    I[\textbf{m}] \mathcal{K}_\theta(\mathbf{\tau})d \mathbf{\tau} \notag\\
    &=
  \sum_{\textbf{m}}I[\textbf{m}]\iint
  \Big(\delta\left(
  \textbf{g}_n-\mathbf{\tau}-\textbf{m}\right)
     \mathcal{K}_\theta(\mathbf{\tau})\Big)d \mathbf{\tau} \notag\\
   &=\sum_{\textbf{m}}
     I[\textbf{m}]\mathcal{K}_\theta(\textbf{g}_n-\textbf{m})
\label{eq:underlying}
\end{align}

where $\textbf{m}=(i,j),  \textbf{n}=(i',j')$ are discrete `pixel' coordinates in $I$ and $I'$ respectively, and $\textbf{g}_n$ denotes the continuous sampling locations of the grid
in Fig.~\ref{fig:underlying} 
(i.e., the sub-`pixel' position of  each `pixel' in $I'$ when projected onto the coordinates of $I$). 

Eq.~\ref{eq:underlying} shows that for $\mathcal{K}_\theta$ with finite support, the output of the continuous convolution at $I'[\textbf{n}]$ is a weighted sum of the discrete ``Neighbors'' of its projected location  $\textbf{g}_n$ in the input $I$. The weights are a function of the distance between the discrete `pixel' center  $\textbf{m}$  to the continuous sampling location~$\textbf{g}_n$.
The desired sampling grid \varbold{$\textbf{g}_n$} is calculated by projecting the output integer `pixel' locations $\textbf{n}=(i',j')$ in $I'$, to sub-`pixel' coordinates in the input $I$. 
We call it the {``Projected Grid''}. 

\begin{figure}[t]
\vspace*{-1cm}
    \centering
    \includegraphics[width=0.9\textwidth]{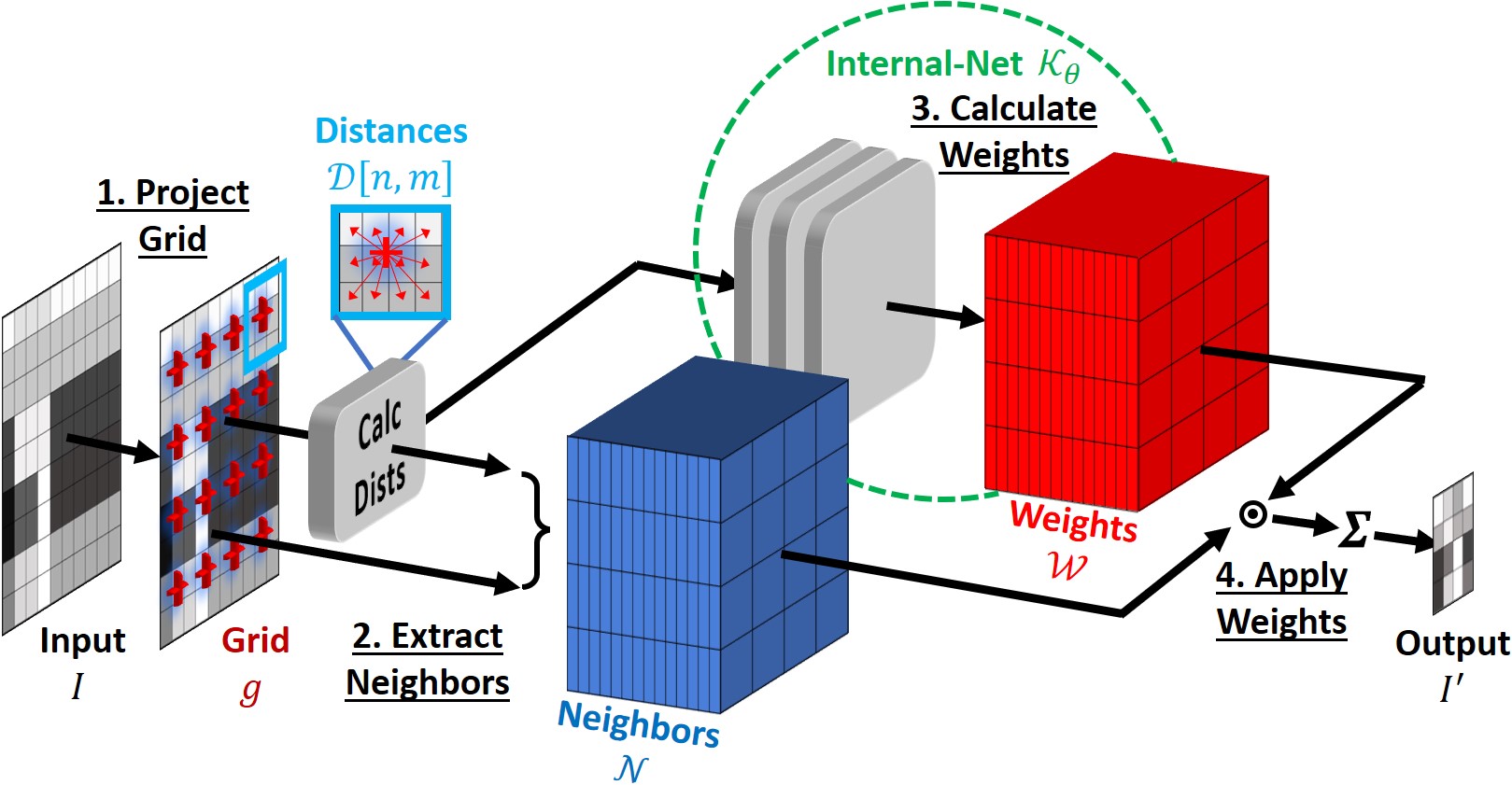}
    \caption{\it Overview of CC-layer. \ (see Sections~\ref{sec:overview} and~\ref{sec:implement} for details)}
    \label{fig:overview}
    \vspace*{-0.3cm}
\end{figure}


In order to execute our approach we need to overcome an inherent challenge:
CNNs are built upon regular grids with discrete coordinates. 
We therefore need to model a continuous process with discrete components. 
Fig.~\ref{fig:overview} illustrates our CC-layer construction, which builds upon implementation ideas from standard image resizing techniques~\cite{MATLAB:2010}. 
Our CC-layer consists of 4 principled building blocks. These are marked by numbers in Fig.~\ref{fig:overview}, and are described in detail in Sec.~\ref{sec:implement}: 
\begin{enumerate}[noitemsep,nolistsep,leftmargin=*]
\item \textbf{Calculate Projected Grid \varbold{$\textbf{g}_n$}} 
from the desired scale-factors and  output-shape. 
\item \textbf{Extract Neighbors \varbold{${\mathcal{N}}$}:} Map each sampling point $\textbf{g}_n$ to its discrete neighbors \varbold{${\mathcal{N}}[\textbf{n}]$} within the continuous kernel support. 
\item \textbf{Calculate Weights} \varbold{$\mathcal{W}$} by applying \textbf{$\mathcal{K}_\theta$} to distances between sampling points to their neighbors:
$\varbold{\mathcal{W}\textbf{}{[n,m]}} = \mathcal{K}_\theta(\textbf{g}_n-\textbf{m})$.
This ``Internal-Net''  \textbf{$\mathcal{K}_\theta$} is the only trainable part in the CC-layer.
\item \textbf{Apply Weights:}  Multiply \varbold{$\mathcal{N} \odot \mathcal{W}$} and sum over all neighbors and input channels, to obtain the output \ben{feature map} $I'$, with any desired scale or shape (which can be determined at test time).
\end{enumerate}

\section{The CC-Layer}
\label{sec:implement}
\subsection{Constructing the CC-layer}
\label{sec:implement_construct}
Given a discrete feature map $\varbold{I}$ of shape $[N,C_{in},H,W]$, we are interested in a layer that produces $\varbold{I'}=CC\{\varbold{I}\}$ of shape $[N,C_{out},H',W']$ which is resized by a (non-integer) scale-factor $(s_h, s_w)$. Here $N$ is the batch-size, $C$ is the number of channels. $H,W$ are spatial dimensions. The desired spatial size $H',W'$ may be chosen at inference time.
Our CC-layer consists of 4 principled building blocks, marked by numbers in Fig.~\ref{fig:overview} (for simplicity, $I$ and $I'$ are drawn as 2D vectors in Fig.~\ref{fig:overview}, i.e.,  4D tensors with $N=1$ and $C=1$). These are described next:

\textbf{Block 1 -- Projected-grid:} 
The Projected grid \varbold{$\textbf{g}$} matches each output `pixel' $\textbf{n}$=$(i', j')$$\in$$I'$ to a sub-`pixel' location \varbold{$\textbf{g}_n$} in the continuous space of the \ben{input} $I$. This grid is captured by a tensor of shape $[2,H',W']$ where the the first dimension are the 2 projected sub-`pixel' coordinates (vertical and horizontal) of each output `pixel' $\textbf{n}$=$(i', j')$$\in$$I'$, and $H', W'$ are for all the output `pixels'.

\begin{figure}
    \centering
    \hspace{-3cm}
    \begin{minipage}{.5\textwidth}
      \centering
      \hspace{1cm}
      \includegraphics[width=1\textwidth]{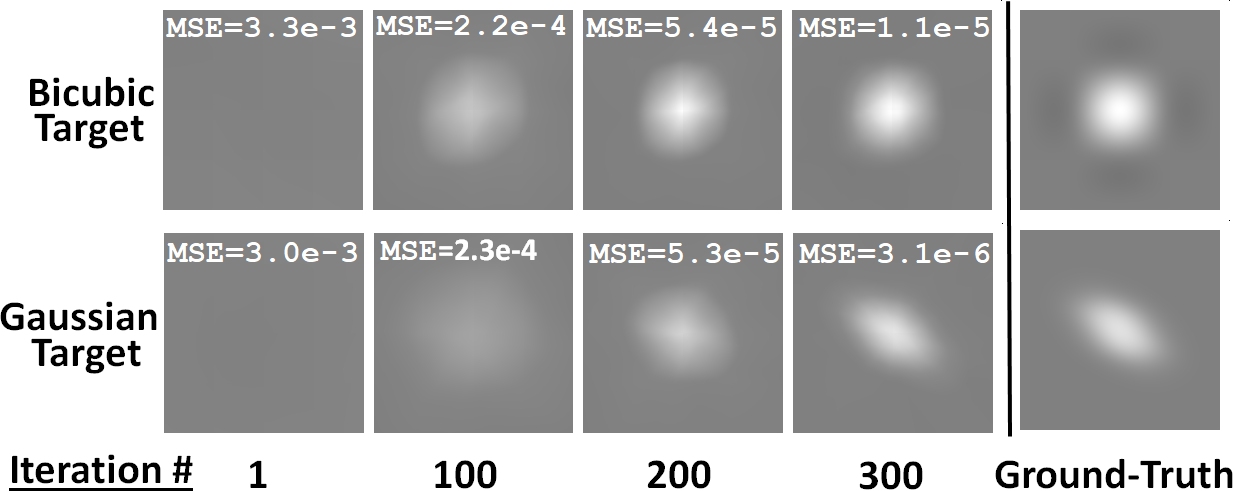}
      \captionsetup{oneside, margin={-0.2cm,0cm}}
      \caption{\mbox{\it Visualizing the continuous learned kernels}}
      \label{fig:visualize}
    \end{minipage}
    \begin{minipage}{.4\textwidth}
        \vspace{-0cm}
        \includegraphics[width=1.5\textwidth]{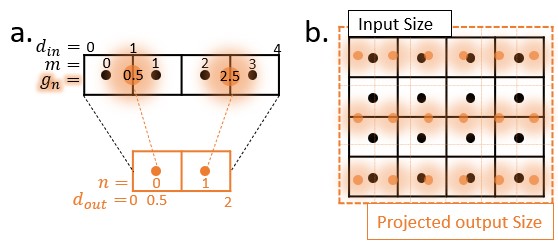}
        \begin{minipage}{1.5\textwidth}
            \captionsetup{oneside, margin={1cm,0cm}}
            \caption{{\it The projected grid.}}
            \label{fig:grid}
         \end{minipage}%
    \end{minipage}%
    \vspace*{-0.5cm}
\end{figure}

Let's start with a simple example. Fig.~\ref{fig:grid}.a depicts a standard 1D  downscaling by 
$2$. Note that even when the scale-factor $s$ is an integer (or an inverse integer: ${s=\ben{\nicefrac{1}{2}}}$ in this case), it is  \emph{wrong} to map an output `pixel' position $\textbf{n}$$\in$$I'$  to its input position in $I$ by simply multiplying (or dividing)  by the scale-factor $s$. It can be observed in Fig.~\ref{fig:grid}.a   that ${\varbold{\textbf{g}_n}\neq2\textbf{n}=\textbf{n}}/{s}$
To obtain the correct mapping, let's
define $d_{out}$  to be the distance of any output `pixel'  
\michal{$\textbf{n}$}
from the leftmost boundary of $I'$.  $d_{out}$  is measured in units of output `pixels'. The matching coordinates \varbold{$\textbf{g}_n$}   in the \ben{input} $I$  is  defined to have a distance $d_{in}$ from the leftmost boundary of  $I$, and is measured in units of input `pixels'. The correct mapping rule across image scales~\cite{MATLAB:2010} is $d_{in}$=${d_{out}}/{s}$, 
since the total shape (from boundary-to-boundary) is resized, rather than discrete `pixel' centers. Since the first `pixel' \emph{center} (in any \ben{feature map/image}) is always half a `pixel' away from the leftmost boundary, hence:
$d_{out}$=$n$+$\frac{1}{2}$  and  $d_{in}$=$\varbold{\textbf{g}_n}$+$\frac{1}{2}$.
This entails the
well known relation used in image resizing methods~\cite{MATLAB:2010}: 
$ \varbold{\textbf{g}_n}$=$\frac{n}{s}$+$\frac{1}{2}$$(\frac{1}{s}$-$1).$

However, we need to handle also non-integer output shapes. The intrinsic output size 
$s \cdot in\_size$ may not be an integer, yet images or feature-maps  can only be represented  by integer-sized vectors/tensors. We will usually (but not always) determine the size of the output $I'$ to be $out\_size =\lceil s \cdot in\_size \rceil$. Fig.~\ref{fig:grid}.b shows an example of a projected grid for resizing a $4 \times 4$ input \ben{feature map} by scales 0.6 and 1.4,  respectively. The final output size ($3 \times 6$) is larger than the intrinsic size ($2.4 \times 5.6$), due to the ceiling operation $\lceil$*$\rceil$.  Therefore, the final  grid mapping (which accurately covers all cases) is:
\begin{align}
    \varbold{g_n} = \frac{n}{s} + \frac{1}{2}\Big( in\_size - 1\Big) - \frac{1}{2s}\Big(out\_size-1\Big)    
    \label{eq:grid}
\end{align}

\textbf{Block 2 -- Neighbors extraction:} For each grid point \varbold{$\textbf{g}_n$}, we now extract all its discrete nearest-neighbors $\varbold{\mathcal{N}[\textbf{n}]}$ \ben{from} $I$. These are all the input `pixels' centers within the support of the continuous kernel $\mathcal{K}_\theta$. This information is captured by a tensor of order 6 (blue tensor in Fig.~\ref{fig:overview}),  with shape $[N,1,C_{in},K,H',W']$ where K is the number of discrete neighbors in the kernel support. The second singleton dimension is for convenience in the next steps. We further need to extract the distances of each sub-`pixel' grid point \varbold{$\textbf{g}_n$} (of an output `pixel' {$\textbf{n}$}),  to all its discrete neighbors  ${\textbf{m}}\in I$. These distances $\varbold{\mathcal{D}[\textbf{n,m}]}$ are kept in a tensor $\varbold{\mathcal{D}}$ (shown in cyan in Fig.~\ref{fig:overview}), whose shape is $[K,2,H',W']$.


\textbf{Block 3 -- Calculate weights:} The weights are produced by a learnable model $\varbold{\mathcal{K}_\theta}$, which is applied to the distances tensor:
$\varbold{\mathcal{W} = \mathcal{K}_\theta \big\{ \mathcal{D}\big\}}$ (similarly to \cite{wang2018deep}).
A weight is assigned to connect between each output `pixel' and all its discrete input neighbors. Therefore, the weight tensor $\varbold{\mathcal{W}}$ will have a shape of $[1,C_{out},C_{in},K,H',W']$ 
(red tensor in Fig.~\ref{fig:overview}). 
Its first singleton dimension is needed to match the size of the Neighbors tensor $\varbold{\mathcal{N}}$. As in standard conv, each output channel  $C_{out}$ is connected to all input channels $C_{in}$ through $\varbold{\mathcal{W}}$. We use a small neural network for $\varbold{\mathcal{K}_\theta}$ called the ``Internal-Net'' (marked in green in Fig.~\ref{fig:overview}). It is a simple CNN sequence of 1$\times$1 conv layers and ReLUs. This way every output `pixel' $\textbf{n}$ connects only to distances $\varbold{\mathcal{D}[\textbf{n}]}$ within its own set of discrete input neighbors. 


\textbf{Block 4 -- Apply weights:} The final stage executes Eq.~\ref{eq:underlying} for the entire output tensor,
by multiplying the Weights tensor (red) with the Neighbors tensor (blue) and summing over all neighbors and input channels:
\vspace*{-0.4cm}
\begin{align}
\varbold{I'}=
CC\{\varbold{I}\}= \sum_{c_{in}, k} \varbold{\mathcal{N}} \otimes \varbold{\mathcal{W}}
\end{align} 
with the following tensor shapes: 

$ \varbold{I'}:[N,C_{out},H',W'], \quad \varbold{\mathcal{N}}: [N,1,C_{in},K,H',W'], \quad \varbold{\mathcal{W}}:[1,C_{out},C_{in},K,H',W'] $

\subsection{Training and Generalization}
\label{sec:train}
CC is end-to-end trainable. The only trainable parameters of CC are $\varbold{\theta}$,  the parameters of $\varbold{\mathcal{K}_\theta}$. The gradient is propagated from the CC output through the weights tensor $\varbold{\mathcal{W}}$ to them. Gradients need to also be propagated to previous layers through the CC input. This is done easily through the neighbors extraction since it is just slicing (each neighbour neuron is actually a copy of an input neuron).

The output shape is determined by the shape of the grid $\varbold{{g}}$. Since $\varbold{\mathcal{K}_\theta}$ is a fully convolutional network, it can be applied to any spatial input size both at training and inference. This means that regardless of what sizes and scales we train on, a single CC, with a single set of parameters $\varbold{\theta}$ is applicable to any size and scale determined at real-time.

The input to  $\varbold{\mathcal{K}_\theta}$ is 
$\varbold{\mathcal{D}}$, which through the grid  $\varbold{{g}}$ depends only on the desired output scale \&
shape. Naturally, to generalize to many scales, we can sample various scales during training. However, being fully 1$\times$1 convolutional, $\varbold{\mathcal{K}_\theta}$ actually maps every 2D distance vector
$\varbold{\mathcal{D}[\textbf{n,m}]}$ to a single value (weight) $\varbold{\mathcal{W}[\textbf{n,m}]}$. This means that $\varbold{\mathcal{D}}$ actually contains a huge batch of
inputs. This produces an interesting advantage: in almost all cases CC generalizes from one scale to any other scale. This happens as long as the diversity of distances in a single grid is reasonable. Eq.~\ref{eq:grid} suggests that if the scale is a rational number with a small numerator, 
then there exists only a small set of grid coordinates, and consequently a small set of unique distances $\varbold{\mathcal{D}[\textbf{n,m}]}$. For example, for $s=\nicefrac{1}{2}$ and a kernel with support of 2$\times$2, we get:
$\varbold{\mathcal{D}[\textbf{n}]} = \Big\{(-\nicefrac{1}{2},-\nicefrac{1}{2}), 
(-\nicefrac{1}{2},\nicefrac{1}{2}), 
(\nicefrac{1}{2},-\nicefrac{1}{2}), 
(\nicefrac{1}{2},\nicefrac{1}{2})
\Big\} \ \ \varbold{\forall \textbf{n}}$, which will not generalize to other scales/distances. In other words, generalization occurs over the distribution of distances $\varbold{\mathcal{D}}$  between the grid points and `pixel' centers. If this distribution collapses to a small set of possibilities, then such generalization is damaged. However, training with a \emph{randomly} selected float scale-factor will give a huge diversity of sub-`pixel' distances in $\varbold{\mathcal{D}}$, hence will be able to generalize with very high probability to any other scale factor.
Empirical evaluation of this property is described in the experiments section and in Fig.~\ref{fig:generalize}.

\section{Features \& Properties}
\label{sec:features}
This section reviews several important and useful properties of CC.
We show that introducing CC-layers gives rise to new CNN capabilities, and furthermore -- allows for \emph{new dynamic architectural designs determined at inference time}.

\textbf{A standard Conv-layer is a special case of the CC-Layer:}
Note that when the scale-factor $s=\nicefrac{1}{k}$, $k \in \mathbb{N}$, and $k$ has the same parity as the kernel support size in both dimensions, $g_n$ in Eq.~\ref{eq:grid} reduces to simple grid locations with integer pixel spacing between them. Thus, all output grid points share the same set of local distances to their discrete input ``Neighbors''. Hence, the set of weights are shared by all output `pixels', which is the case in standard convolution. This implies that  for integer scale ratios CC reduces to standard convolution, i.e.,  CC is a \emph{generalization} of the standard Conv-layer.


\textbf{Dynamic scale-factor at inference:} As mentioned, the projected-grid is purely a function of the scale-factors and output shape. It is also important to note that in case of training the CC-layer with a \emph{random float scale factor}, we obtain a huge diversity of distances in $\varbold{\mathcal{D}}$ (see more details in Sec.~\ref{sec:train}). Hence the ``Internal-Net'' $\varbold{\mathcal{K}_\theta}$ which gets all the sub-`pixel' distances $\varbold{\mathcal{D}}$ as an input, and outputs a single kernel  consistent with them all,  must learn a  \emph{true continuous} weight function (the continuous kernel), independently of the scale factor or shape of the \ben{output}. Once it has recovered the continuous kernel (continuous weights),  CC can be applied at inference time to output any dynamically chosen scale or shape, simply by changing the sampling grid.


\textbf{True Shift-Invariance/equivariance:} It was shown~\cite{zhang2019shiftinvar, azulay2018deep} that strided conv-layers \niv{lack}
equivariance to shifts of the \ben{input}. This mostly happens due to aliasing. The common filter size tends to be smaller than the low-pass filter size required to remove high-frequencies when downscaling by a factor of 2. CC allows more gradual downscaling, so that (with the same kernel support) the sampling frequency is higher and less aliasing occurs. For example, one can replace a single strided convolution layer with downscaling-factor $s=1/2$, with a few (2-3) CC-layers, each with scale $1/2 < s < 1$ (e.g., 2 CC-layers with $s=1/\sqrt{{2}}$, or 3 CC-layers with $s=1/\sqrt[3]{{2}}$).

\textbf{Standard conv-layers often suffer from inherent misalignments, which are ameliorated by CC-layers:} Examining the accurate grid mapping in Eq.~\ref{eq:grid}, one can see that there are cases in standard discrete convolutions which fail to satisfy it. Such cases occur when the size of the filter and the stride have different parities, in some dimension. The intuitive reason is that the filter \emph{center} is defined by its parity: In odd-sized filters, the center of the output `pixel' falls on an input `pixel'-center, whereas in even-sized filter, it falls on the boundary between two input `pixels'. As mentioned, for a scale factor $1/int$, the set of weights is identical for any output position. Eq.~\ref{eq:grid} suggests that in such cases the grid locations are either all exactly on input `pixel' centers, or all on the boundary between input `pixels' (e.g., scale-factor of $\frac{1}{2}$ produces locations 0.5, 2.5, 4.5 etc., whereas scale factor of $\frac{1}{3}$ produces locations 1,4,7 etc.) This means that in case of an even scale with odd-sized filter, standard conv-layers introduce small misalignments between the input and output ($I$,$I'$). These small \benc{del: layer-to-layer} misalignments can accumulate to large misalignments over many layers. 

\begin{figure}
\hspace{-0cm}
\centering
\begin{minipage}{.45\textwidth}
    \hspace{-1cm}
    \includegraphics[width=1\textwidth]{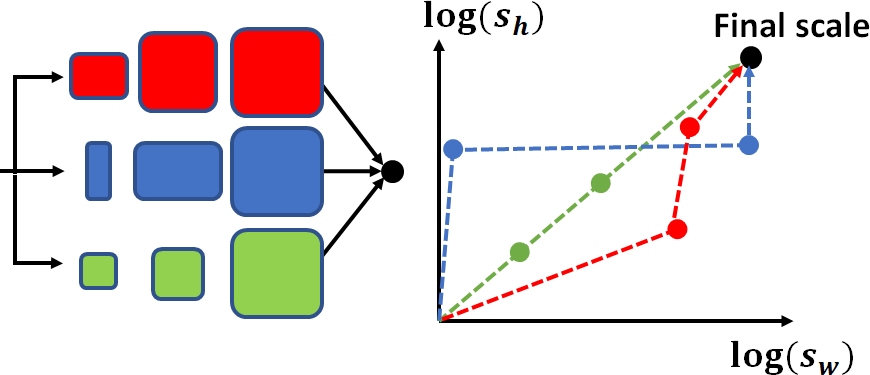}
    \captionsetup{oneside, margin={-2cm,0cm}}
     \caption{\it ``Scale-Ensemble'' (see Sec.~\ref{sec:features}  for details).}
    \label{fig:scale_ensemble}
\end{minipage}%
  \hspace{0.3cm}
  \begin{minipage}{.3\textwidth}
  \centering
  \includegraphics[width=1\textwidth]{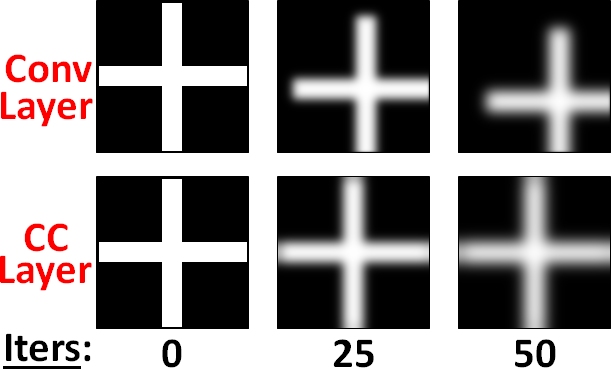}
   \captionsetup{oneside, margin={-0.5cm,-1cm}}
  \caption{\it Standard conv-layers induce misalignments.  CC-layers do not.}
  \label{fig:fig_missalign}
    \end{minipage}
   \vspace*{-0.5cm}
\end{figure}

Such an example is shown in Fig.~\ref{fig:fig_missalign}. 
We used a simple gaussian kernel,  and applied it repeatedly to an input image (white cross), using either a sequence of standard conv-layers or a sequence of CC-layers, all with stride=1 or scale=1, respectively. We chose an even-sized 4$\times$4 kernel. It can be seen that standard conv-layer shifts the result by half a pixel per iteration, resulting in a large shift after many iterations. In contrast, CC-based convolutions remain centered, due to the accurate sub-pixel grid mapping of Eq.~\ref{eq:grid}. Choosing a different padding method for the standard conv would only result in shifts to other directions; there is no way to apply a 4$\times$4 discrete convolution that keeps the same input size and does not induce misalignments. The reason neural networks still get good results is simple: they learn a variety of shifted kernels. This misalignment phenomenon has been observed in super-resolution works~\cite{zhang2020deep, ZSSR}. Still, the way  misalignments are currently addressed is non-ideal and unprincipled, as some portion of the filter size is not used and may create artifacts.

\textbf{Scale Ensembles:} Here we present \niv{an entirely}
new capability in Deep-Learning -- Dynamic Architectural Design of CNNs, at inference time, which is made possible by introducing multiple CC-layers into CNNs.  As mentioned before, a trained CC-layer can generate any output scale-factor chosen at inference time. Therefore, a single trained CNN which consists of multiple CC-layers, can be applied many times at inference, each time with \emph{a different sequence of scales}, as long as the final output scale \& shape of the entire network remain fixed. This means that there are infinitely many ways to apply such a trained network, and still get the same final output size/scale.  Moreover, each dimension (vertical or horizontal) can be scaled differently at intermediate CC-layers. This gives rise to a new type of ``self-ensembles'' of CNNs. We call it ``Scale Ensembles''.  Fig.~\ref{fig:scale_ensemble} schematically illustrates a few different scales-sequences for 3 consecutive CC-layers in the same pre-trained CNN. For example, the Green sequence applies 3 consecutive uniform scaling in both dimensions, same scaling each time; 
the Blue sequence scales  the horizontal and vertical dimensions alternatingly; whereas the Red sequence offers yet another non-uniform sequence of scaling. At inference time we can apply the same pre-trained network several times, each with a different sequence of scales. We can then aggregate all the results of this ensemble,  to get an improved result  which is consistent with the entire ensemble. Aggregation can be achieved, e.g., by  averaging the class \niv{logits}
in a classification task,  or by computing the median image of the ensemble in an image-processing task.

\section{Efficient Implementation}
\vspace{-0.3cm}
\label{sec:efficiency}
This section provides major implementation details which give rise to efficient implementation of the CC-layer. Code will be made available.

\textbf{No need to keep the Neighbors distances:} The distances $\varbold{\mathcal{D}}$  of a grid point  \varbold{$\textbf{g}_n$}  to all its discrete input Neighbors can actually be predicted by keeping only one Neighbor. The reason is simple: all neighbors are at a sequence of positions 1 `pixel' apart from each other. 
It suffices to keep only the distance of a grid point to its \emph{closest} `pixel' center in order to predict all other distances. Moreover, the distance to each Neighbor is already implied by the sub-`pixel' grid coordinates. Hence, we can train the Internal-Net
$\varbold{\mathcal{K}_\theta}$  to predict the weights \emph{directly from the grid} \varbold{$\textbf{g}$}, thus saving on memory. 


\textbf{Efficient implementation using discrete convolutions:} Eq.~\ref{eq:grid} 
suggests that special types of grids exist. We already noted that for $1/int$ scales, CC collapses to a standard convolution. However, Eq.~\ref{eq:grid}  further suggests that for \emph{rational-numbers}  scale-factors ($s=\nicefrac{k}{\ell}$) the grid is periodic, with a period equals the numerator $k$. This generalizes observations made by~\cite{romano2016raisr,freedman2011image}. We take advantage of this property by calculating the sets of weights for one period only, to avoid calculation of a huge weights tensor $\mathcal{W}$. We do not need to explicitly extract neighbors and multiply them with the weights. Instead, we use a set of standard discrete convolution filters, each with a different starting shift matching a different grid position within one period. We apply them to the input feature-map and interleave their results. This allows us to use the existing, highly optimized, implementation of conv operation. This approach offers a trade-off between speed (small $k$) and better generalization (large $k$). \nivc{the unattentive reviewer might understand from this section that all we do is use a grid of convs instead of one conv. we need to somehow emphasize that this grid can principally change at each iteration, unlike ordinary conv-grids}


\textbf{Saving memory by keeping only $\mathcal{K}_\theta$:} The most memory consuming tensor in CC is the weights tensor $\mathcal{W}$. The second one is the Neighbors tensor $\mathcal{N}$. We can save almost all the memory used by a CC-layer by simply removing these tensors at each iteration after the output is calculated. In the back-propagation step we simply recalculate them, \ben{at some extra cost of time. The memory footprint can be further reduced by chunking the input and computing the per-chunk results sequentially}.
\section{Experiments}
\vspace{-0.3cm}
\label{sec:applications}

\textbf{Visualizing the continuous learned kernels $K_\theta$:} Filters of discrete conv-layers can be easily visualized. They are sets of discrete numbers on a regular grid, which can just be printed out or viewed as images. Filters of CC-layers, on the other hand, assign a unique set of numbers to any sub-`pixel' grid location. But once trained,
we can easily sample $K_\theta$ on a regular grid at any desired resolution for the purpose of visualization. Fig.~\ref{fig:visualize} shows 2 such examples of CC kernel visualization. In both examples we trained a single CC-layer to \emph{imitate} an image-resizing method (for which we know the ground-truth kernels): (i) Imitate bicubic resizing, and (ii) imitate resizing by a Gaussian kernel with different $\sigma$ for each axis, rotated by~\ang{45}.  The CC-training was performed  on a single image with a single random sampled scale $s$$\in$$[0.3$$,..,$$1.3]$ in each dimension. The input to the CC-layer was the single image, and the ``output label'' was the resized image  (generated by the external image-resizing method for that $s$).  
Once trained, we sampled $K_\theta$ at dense 200$\times$200 coordinate-pairs.  Fig.~\ref{fig:visualize} shows such visualizations of the  continuous recovered kernel $K_\theta$,  as a function of the training iteration.

\textbf{Verifying  scale generalization:} To test generalization across scales, we sampled a \emph{single}  pair of random scale-factors $s_x$,$s_y$$\in$$[0.3$$,..,$$1.3]$, and applied an independent bicubic resizing to a single image for this scale. As before, we trained a CC-layer with the original image as  the input, and the resized image as the ``output label''. To test generalization, we applied our CC-layer  (which was {trained} for one scale), on 100 \emph{new random scales}. We compared the resulting 100 output images to ground-truth bicubic resizing by those 100 scales. The left side of Fig.~\ref{fig:generalize} shows  MSE of the train-error and the generalization (test) error. The test-error for scales CC never trained for, is very close to the train-error for the single scale it trained on. This is due to the fact that CC generalizes well to other scales, as long as the training scale generates sufficient diversity in sub-`pixel' grid shifts. To further confirm this, we repeated the experiment, only this time we chose the training scale to be $1/int$ (1/2). Eq.~\ref{eq:grid} shows that in such case all grid locations have the same shift from `pixel' centers (0.5). The right side of Fig.~\ref{fig:generalize} shows that generalization was severely damaged; test-error for 100 other scales was 2-orders-of-magnitude higher. Fig.~\ref{fig:generalize} further shows visualization of $\mathcal{K}_\theta$ for these 2 experiments (ground-truth can be found in Fig.~\ref{fig:visualize}). In the first experiment the learned kernel is similar to the ground-truth, whereas in the second experiment it
is not (since it was only required to produce correct results for a very small subset of coordinate pairs).

\textbf{Verifying Shift-Equivariance:}
To check shift-equivariance, we used a variant of a very simple CNN-based classification net for CIFAR-10~\cite{CIFAR10} (similar to the net used in~\cite{wong2018scaling, atzmon2019controlling}). The network consisted of several conv-layers of strides 1,2,1,1,1,1,1,2, followed by 3 fully-connected layers, all with ReLU activations. We call this the ``Baseline Net''.
We constructed a ``CC-Net'' 
with the same number of layers, but with CC-layers of \emph{gradually} changing layer sizes (starting from the second layer). The overall scale factor from the net's input to the last conv-layer's output remained the same (1:4), and the scale-factors of all layers were approximately ${(\nicefrac{1}{4})^{\nicefrac{1}{6}}}$. When testing on the CIFAR-10 test-set, CC-Net had higher accuracy (87.81\%) compared to the Baseline (87.14\%), which is not surprising as CC-Net has more params than Baseline-Net, but was not the purpose of this experiment. 

\niv{Following}~\cite{zhang2019shiftinvar}, we tested the shift-equivariance as follows: (i)~Each input image was fed to the network twice: First the original input, and then, a shifted version of it with horizontal/vertical shifts up to a max allowed radius. (ii)~We took the output of the last conv-layer. Since it is 4 times smaller, a 1-pixel shift in the input image should translate to 0.25-`pixel' shift in the output feature-map of the last layer. We therefore `shift-back' the output feature-map by the expected shift (using cubic-interpolation), to bring it back to alignment with the original (unshifted) output.
(iii)~We compute the cosine similarity between the 2 output feature-maps after alignment. To avoid boundary effects, we cut off 2 boundary `pixels' from each side.
%
\niv{The results are shown in Table.~\ref{table:equivar}.} We tested it for 2 types of inputs: images \& Gaussian noise. The table shows that CC-layers are much more resilient to shifts than regular conv-layers (both on natural images and on random noise). 

\begin{minipage}{1\textwidth}
\begin{minipage}[b]{.4\textwidth}
\centering
    \begin{tabular}{|c||c|c|c|}
    \hline
                                    & Max-shift   &{Baseline}       & {CC-net}           \\ \hline\hline
    \multirow{3}{*}{CIFAR-10}  & $\pm 2$       & 0.90            & \textbf{0.92}          \\ \cline{2-4}                  
                                   & $\pm 4$       & 0.78         & \textbf{0.85}          \\ \cline{2-4}
                                    & $\pm 6$       & 0.65        & \textbf{0.77}          \\ \hline
    \multirow{3}{*}{Noise}          & $\pm 2$       & 0.80        & \textbf{0.91}          \\ \cline{2-4}
                                   & $\pm 4$       & 0.67         & \textbf{0.85}          \\ \cline{2-4}
                                  & $\pm 6$       & 0.56          & \textbf{0.85}          \\ \hline          

\end{tabular}  
   \captionsetup{oneside, margin={0.5cm,-1cm}}
\captionof{table}{Comparison of shift equivariance results, measured in cosine-similarity}
\label{table:equivar}
\end{minipage}%
\hspace{3cm}
  \begin{minipage}[b]{0.35\textwidth}
  \includegraphics[width=1.1\textwidth]{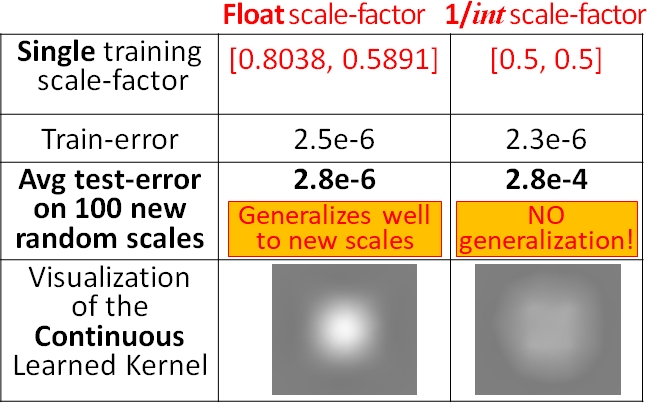}
   \captionsetup{oneside, margin={-0.1cm,-0.2cm}}
  \captionof{figure}{\it  CC trained on one (float) scale, generalizes well to other scales.}
    \label{fig:generalize}
\end{minipage}

\end{minipage}

\textbf{The power of dynamic Scale-Ensembles at inference:}
Since \michal{the above} ``CC-Net'' was trained with non-integer (float) \michal{internal} scale-factors, it can generalize to new \michal{internal} scale-factors \emph{at inference-time}. This gives rise to \emph{ensemble-based image classification} (Sec.~\ref{sec:features} \& Fig.~\ref{fig:scale_ensemble}). 
This means that our trained CC-Net can be applied numerous times at inference, each time with a different sequence of scales (as long as their product yields 1/4), and average the \niv{logits}
of all forwarded passes.
\michal{We tested this for CC-Net. Even small ensembles (2,3)}
already gave a significant boost in classification accuracy (+1.7\%, +2.4\%, respectively).
\michal{While test-time augmentations for regular CNNs can be applied  to the input image only, Scale-Ensembles apply test-time augmentation to \emph{all} CC-layers in the network. Hence, deep nets with many CC-layers, are likely to benefit from larger self-ensembles.}
%

\textbf{Potential use in image-processing tasks:}
CC-layers could potentially be very useful in image-processing tasks. For example, existing Super-Resolution (SR) networks are trained for a fixed scale-factor~\cite{EDSR, srcnn}. To increase the resolution of an image by a factor of 3, one must use a SRx3 net which was trained for this specific scale-factor. SR by a factor of 4 will require a different pretrained SRx4  net. 
Constructing a SR network based on CC-layers is expected to give \emph{SR with any desired scale factor} (or at least an impressive range of scale-factors), all \emph{with a single trained SR net}. 
\michal{Moreover, it could potentially accommodate different SR scale factors for different axes (if so desired). Note that training a separate net for every potential SR scale 
(or any combination of vertical~x~horizontal SR scales) 
is combinatorially daunting. A CC-based SR network may alleviate this problem.}
Designing such a CC-based SR network is part of our future work.

\clearpage
\printbibliography

\clearpage
\begin{appendices}
\section{Details of Experiments and Architectures}

\subsection{Internal Net $\mathcal{K}_\theta$}
Table.~\ref{tab:internal} below shows the architecture of the internal net $\mathcal{K}_\theta$ used for all experiments. The properties of each layer is denoted by  \textsc{Conv} $[Channels], [Kernel\_height] \times [Kernel\_width] + [Stride]$. Each Conv layer of  $\mathcal{K}_\theta$ except the last one, is followed by LeakyReLU activation. Each \CC layer contains its own \textsc{InternalNet}$: \mathbb{R}^2 \to \mathbb{R}$ (the learned continuous kernel). The input to \textsc{InternalNet} are distances from a specific pixel position in the output, given as a 4D tensor, and computed using $3$ consecutive 2D 1$\times$1 convolutions, as detailed in the table .

Block 2 in Sec.~\ref{sec:implement_construct} describes the input to $\mathcal{K}_\theta$ -- distances tensor of size $[K,2,H',W']$ where K is the number of neighbors within the support. When applying 2D conv layers to it, the first dimension of the tensor is regarded as the batch dimension. We do so intentionally, as we want to apply the same operation to all neighbor distances in parallel. The final calculated weight tensor is of size $[1,C_{out},C_{in},K,H',W']$, where $C_{in}, C_{out}$ are the numbers of input channels and output channels, respectively. 

By default, the number of channels of the final internal conv layer,  {$C_{final}$} needs to cover both in and out external channels of the CC layer. {$C_{final} = C_{in} \times C_{out}$}, then the calculated weights tensor  can be obtained by simple reshaping of the Internal Net's output: 
$[K, C_{out}\times C_{in}, H', W'] \rightarrow [1,C_{out},C_{in},K,H',W']$. 
In the equivariance experiment we used the efficient implementation mentioned in Sec.~\ref{sec:efficiency}: "No need to keep the Neighbors distances". In this case, the input is just the grid [1,2,H',W'] so the batch size is 1 and {$C_{final} = C_{in} \times C_{out} \times K$} to accommodate for the weights of all the neighbors.

\begin{table}[h!]
    \centering
    \begin{tabular}{lll}
    \toprule
       InternalNet  \\
    \midrule
      \textsc{Conv} 16 1x1+1  \\
      \textsc{Conv} 16 1x1+1  \\
      \textsc{Conv} $C_{final}$ 1x1+1  \\
    \bottomrule
    \end{tabular}
    \vspace{5pt}
    \caption{Architecture used for $\mathcal{K}_\theta$ in all experiments}
    \label{tab:internal}
\end{table}

\subsection{Shift Equivariance Experiments}
All experiments were conducted on \textsc{CIFAR10} dataset \cite{CIFAR10} and implemented using \textsc{PyTorch}~\cite{paszke2017automatic} deep learning framework. 

The architectures we used for the Classification Networks in Sec.~\ref{sec:efficiency} are detailed in Table~\ref{tab:architectures}. \textsc{CC} corresponds to a continuous convolution layer with similar terminology (except that \textsc{CC} layers do not have a fixed stride). \textsc{FC} $n$ corresponds to a fully connected layer with $n$ outputs. Each \textsc{Conv}/\textsc{FC} layer is followed by a ReLU activation except for the last fully connected layer. As can be seen, \textsc{CCNet} and \textsc{ConvNet} share the same general architecture. The models are based on similar models from \cite{wong2018scaling,atzmon2019controlling}, where we added a few more convolution layers to exemplify the gradual change in scale.

\begin{table}[h!]
    \centering
    \begin{tabular}{lll}
    \toprule
         \textsc{ConvNet (Baseline)}  & \textsc{CCNet}\\
    \midrule
         \textsc{Conv} 32 3x3+1     & \CC 32 3x3      \\
         \textsc{Conv} 32 4x4+2     & \CC 32 4x4      \\
         \textsc{Conv} 64 3x3+1     & \CC 64 3x3      \\
         \textsc{Conv} 64 3x3+1     & \CC 64 3x3      \\
         \textsc{Conv} 64 3x3+1     & \CC 64 3x3      \\
         \textsc{Conv} 64 3x3+1     & \CC 64 3x3      \\
         \textsc{Conv} 64 3x3+1     & \CC 64 3x3      \\
         \textsc{Conv} 64 4x4+2     & \CC 64 4x4      \\
         \textsc{FC} 512            & \textsc{FC} 512 \\
         \textsc{FC} 512            & \textsc{FC} 512 \\
         \textsc{FC} 10             & \textsc{FC} 10  \\
    \bottomrule\\
    \end{tabular}
    \vspace{-5pt}
    \caption{Model architectures used for shift-equivariance experiment}
    \label{tab:architectures}
\end{table}

\paragraph{Initialization} The outputs of the \textsc{InternalNet} are equivalent to the weights of an ordinary \textsc{Conv} layer. At initialization, we require those outputs to have a variance similar to an ordinary \textsc{Conv} layer's weights at initialization (either \textit{Xavier} \cite{glorot2010understanding}, \textit{Kaiming} \cite{he2015delving} or the default \textsc{PyTorch}~\cite{paszke2017automatic} initializations). This is done by initializing the biases of the last layer of the \textsc{InternalNet} using a normal distribution with the required variance.

\paragraph{Training Hyperparameters} All models were trained using a batch size of $64$ of images obtained by padding the original CIFAR10 images by 8 rows/cols (4 from each side) and taking a random crop of $32 \times 32$. We also use
random horizontal-flip and normalize the inputs to have zero mean and unit variance (as done with common models for training on \textsc{CIFAR10}~\cite{pytorch-cifar}). 
We used \textsc{ADAM} optimizer \cite{kingma2014adam} with weight-decay of $0.0005$. For \textsc{CCNet} we use learning rate of $0.0005$ with a reduction by $10$ on epochs $[200,300]$. For the baseline \textsc{ConvNet} model we checked learning rates $0.0001, 0.0005, 0.001, 0.002$ with several learning rate scheduling (reducing by $10$) at: $[70,150], [150,250], [100,250], [200,300]$, with and without weight-decay, and chose the best performing model of learning rate $0.001$, weight-decay of $0.0005$ and learning rate reduction at epochs $[150,250]$. Results are reported after $400$ epochs, where all our runs seem to fully converge (e.g., further reduction in learning rates did not change the reported results).

\paragraph{Training and Inference with scale-augmentations} 
As mentioned, \CC layers can be given the scale and output shape dynamically during forward pass. In order to train a \textsc{CCNet} model with scale-augmentations we use the following design: the first \CC layer has a fixed scale of $1$ and output shape equal to input shape. We are left with $7$ \CC layers to reduce spatial dimension by $1/4$, that is, we want the product of all $7$ \CC layers' scales to equal to $1/4$. The \CC layers' scales are then sampled from a normal distribution of mean $(1/4)^{(1/7)}$ and standard deviation of $0.01$. All scales are then projected to the nearest rational fraction with a maximal denominator of $10$. At this point we have $7$ scales $\{s_i \}_{i=1}^7$, however $\Pi s_i$ does not necessarily equal to our target scale $1/4$. To this end we uniformly at random choose one layer $j$ and set $s_j = (1/4) / \Pi_{i \neq j} s_i$. At inference time we sample $k$ scales using the same methodology described here. The output shapes are determined by the following:
\begin{equation}
    \mathrm{output\_shape}\left(j\right) = \begin{cases} 
          \Bigl\lceil{\mathrm{input\_shape} \times \Pi_{i=1}^j s_i }\Bigr\rceil & j\mathrm{\ is\ not\ last} \\
          \mathrm{input\_shape} \times (1/4) & j \mathrm{\ is\ last\ layer} \\
   \end{cases}
   \label{eq:output_shape}
\end{equation}

A few examples for scale augmentations are shown in Table~\ref{tab:scale_augs_examples}.

\renewcommand{\arraystretch}{1.3}
\begin{table}[h!]
    \centering
    \begin{tabular}{l|cc|cc}
    \toprule
                & \multicolumn{2}{c}{Augmentation Example 1}                     & \multicolumn{2}{c}{Augmentation Example 2}                        \\
         Layer  &                {Scale [H,W]}                                & $\to$ Output Shape   & {Scale [H,W]} & $\to$ Output Shape         \\
    \midrule
        $1$     &                $[ \frac{1}{1}$    , $\frac{1}{1}]$          &   $[32,32]$          & $[\frac{1}{1}$  , $\frac{1}{1}]$     & $[32,32]$      \\        
        $2$     &                $[ \frac{5}{6}$    , $\frac{7}{9}]$          &   $[27,25]$          & $[\frac{75}{98}$, $\frac{7}{8}]$     & $[25,28]$      \\        
        $3$     &                $[ \frac{5}{6}$    , $\frac{4}{5}]$          &   $[23,20]$          & $[\frac{4}{5}$  , $\frac{2}{3}]$     & $[20,19]$      \\        
        $4$     &                $[ \frac{8}{9}$    , $\frac{5}{6}]$          &   $[20,17]$          & $[\frac{4}{5}$  , $\frac{3}{4}]$     & $[16,14]$      \\        
        $5$     &                $[ \frac{189}{250}$, $\frac{3}{4}]$          &   $[15,13]$          & $[\frac{7}{8}$  , $\frac{5}{6}]$     & $[14,12]$      \\        
        $6$     &                $[ \frac{6}{7}$    , $\frac{6}{7}]$          &   $[13,11]$          & $[\frac{5}{6}$  , $\frac{5}{6}]$     & $[12,10]$      \\        
        $7$     &                $[ \frac{5}{6}$    , $\frac{7}{8}]$          &   $[11,10]$          & $[\frac{7}{9}$  , $\frac{864}{875}]$ & $[9\ ,10]$      \\        
        $8$     &                $[ \frac{3}{4}$    , $\frac{6}{7}]$          &   $[8\ ,\ 8]$        & $[\frac{9}{10}$ , $\frac{5}{6}]$     & $[8\ ,\ 8]$    \\
    \bottomrule
    \end{tabular}
    \vspace{5pt}
    \caption{Examples of scale augmentations. $[H,W]$ corresponds to each spatial dimension. The model consists of $8$ continuous convolution layers (where the first is fixed to a scale $[1,1]$), input-shape is $[32,32]$ and target-scale is $[\nicefrac{1}{4},\nicefrac{1}{4}]$. The output shapes are computed from the scales and input shape using Equation~\ref{eq:output_shape}.}
    \label{tab:scale_augs_examples}
\end{table}


\section{Runtime and Memory Analysis}
\subsection{CC Layers}
We measured the runtime and memory consumption of CC layers, with varying input size and number of consecutive layers. We show that our efficient implementation techniques (introduced in Sec.~\ref{sec:efficiency}) significantly reduce the memory footprint of a CC-layer, and some techniques (implementation using discrete convolutions) further speed up the layer. For reference, we also compare to a Conv-layer, although it lacks the expressiveness of the CC-layer. In the column ``CC (clear memory)'' (see Tables~\ref{tab:runtime_by_sz} and~\ref{tab:runtime_by_num_layers}), we divided the computation into chunks of size $32\times32$, computed the results sequentially and discarded all intermediate results (see "Saving memory by keeping only $\mathcal{K}_\theta$" in Sec.~\ref{sec:efficiency}). This implementation supports non-rational scales, and allows us to use deeper networks than the non-optimized implementation (``CC (standard)'' column). In the column ``CC (by discrete conv)'', the number of computed filters is reduced from the output size to the numerator of the scale factor -- in our case the scale factor is $\nicefrac{2}{3}$ so we have 2 filters. This mechanism is further optimized by using the convolution operation as backend (see "Efficient implementation using discrete convolutions" in Sec.~\ref{sec:efficiency}). This implementation significantly reduces memory consumption and runtime, at the cost of expressiveness -- as it only supports rational scale factors (and is beneficial as long as the numerator is small).

In terms of runtime, we measured the time of a single forward and backward pass through the layer (or layers). To accurately measure the time, we run 11 repeats and excluded the first (warm-up), and report the average of the last 10 (the deviation was negligible). In terms of memory consumption, we report the peak allocated memory in GiB ($2^{30}$ bytes). Sometimes there is a difference between the peak reserved memory and the peak allocated memory, but that's more related to the internals of \textsc{PyTorch}~\cite{paszke2017automatic}. All the tests were conducted on a single Tesla V100-PCIE-16GB GPU.

Table.~\ref{tab:runtime_by_sz} shows the runtime and memory consumption of a single layer with varying input size. In all the experiments we used a $3\times3$ kernel with 32 input channels and 32 output channels, and a batch size of 50. For the CC-layer, we used inner architecture as in Table.~\ref{tab:internal} with scale of $\nicefrac{2}{3}$. For the Conv-layer, we used stride of 1.

Table.~\ref{tab:runtime_by_num_layers} shows the runtime and memory consumption of multiple layers with a fixed input size of $64\times64$. In this experiment we used the same parameters, except for the scale -- here we used a scale of $1$ to have the same input size in all layers.

\begin{table}[h!]
    \centering
    \begin{tabular}{c|cccc}
    \toprule

Input Size     & CC (standard)     & CC (clear memory) & CC (by discrete convs) & Conv (reference) \\
\midrule
$32\times32$   & 8.7ms / 0.9GiB    & 19.7ms / 1.8GiB   & 3.5ms / 0.1GiB         & 0.8ms / 0.1GiB   \\
$64\times64$   & 30.0ms / 3.4GiB   & 76.1ms / 3.7GiB   & 3.6ms / 0.2GiB         & 1.3ms / 0.2GiB   \\
$96\times96$   & 72.3ms / 7.6GiB   & 155.4ms / 3.9GiB  & 5.1ms / 0.4GiB         & 2.7ms / 0.5GiB   \\
$128\times128$ & 131.5ms / 13.6GiB & 285.9ms / 4.1GiB  & 7.6ms / 0.7GiB         & 4.0ms / 0.4GiB   \\
    \bottomrule

    \end{tabular}
    \vspace{5pt}
    \caption{Runtime and memory consumption of CC-layer with varying input size.}
    \label{tab:runtime_by_sz}
\end{table}

\begin{table}[h!]
    \centering
    \begin{tabular}{c|cccc}
    \toprule
\# Layers & CC (standard)   & CC (clear memory) & CC (by discrete convs) & Conv (reference) \\
\midrule
1        & 71.7ms / 7.5GiB & 158.7ms / 3.8GiB  & 3.5ms / 0.3GiB         & 1.3ms / 0.2GiB   \\
2        & OUT OF MEMORY   & 317.9ms / 3.9GiB  & 7.0ms / 0.3GiB         & 2.8ms / 0.3GiB   \\
4        & OUT OF MEMORY   & 635.0ms / 4.0GiB  & 14.1ms / 0.4GiB        & 6.0ms / 0.4GiB   \\
8        & OUT OF MEMORY   & 1270.2ms / 4.2GiB & 27.1ms / 0.7GiB        & 12.5ms / 0.6GiB  \\
16       & OUT OF MEMORY   & 2538.7ms / 4.6GiB & 54.6ms / 1.1GiB        & 25.2ms / 1.0GiB  \\
32       & OUT OF MEMORY   & 5074.7ms / 5.5GiB & 111.3ms / 2.0GiB       & 51.6ms / 1.8GiB  \\
    \bottomrule
    \end{tabular}
    \vspace{5pt}
    \caption{Runtime and memory consumption of varying number of consecutive CC-layers.}
    \label{tab:runtime_by_num_layers}
\end{table}

\subsection{CC Classification Networks}
\textsc{Baseline ConvNet} takes 6 seconds per epoch, while \textsc{CCNet} takes 90 seconds per epoch. This difference is mainly attributed to the fact that CC-layer has a constant overhead of computing the weights, which is independent of the spatial size (when using conv implementation) and thus more significant when the input is small. Also, the feature maps in the \textsc{CCNet} are always (spatially) larger than in \textsc{Baseline ConvNet} due to the gradual downscaling. Finally, some scale factors (which were randomly chosen at each iteration) has large numerator which reduces the speed of the conv-based implementation.

\end{appendices}

\end{document}